\documentclass[runningheads]{llncs}

\usepackage[mobile]{eccv}

\usepackage{eccvabbrv}

\usepackage{graphicx}
\usepackage{booktabs}

\usepackage[accsupp]{axessibility}  % Improves PDF readability for those with disabilities.

\usepackage{hyperref}

\usepackage{orcidlink}

\begin{document}

\title{ColorwAI: Generative Colorways of Textiles through GAN and Diffusion Disentanglement} 

\titlerunning{ColorwAI: Disentanglement Colorway Creation of Textiles}

\author{Ludovica Schaerf\inst{1}\orcidlink{0000-0001-9460-702X} \and
Andrea Alfarano\inst{1}\orcidlink{0009-0002-8031-6103} \and
Eric Postma\inst{2}\orcidlink{0000-0001-9627-1523}}

%\author{Ludovica Schaerf\inst{1,2}\orcidlink{...} \and
%Andrea Alfarano\inst{3}\orcidlink{...} \and
%Eric Postma\inst{4,5}\orcidlink{...}}

\authorrunning{L.~Schaerf et al.}

\institute{Institute for Digital Visual Studies, University of Zurich - Max Plank Society, Culmannstrasse 1, 8006 Zurich, Switzerland \and Cognitive Science \& AI, Tilburg University, The Netherlands \\
\email{ludovica.schaerf@uzh.ch}\\
 }
\maketitle

\begin{abstract}
  Colorway creation is the task of generating textile samples in alternate color variations maintaining an underlying pattern. Selecting a colorway is a complex creative task, responding to client and market needs, technical and cultural specifications, and personal artist style. We introduce a framework, "ColorwAI", to tackle the generative task using color disentanglement on StyleGAN and Diffusion while maintaining minimal shape alteration. We present a variation of the InterfaceGAN method for semi-supervised disentanglement, ShapleyVec, which uses Shapley values to subselect salient dimensions from the detected latent direction. Moreover, we present a framework to employ common disentanglement methods on any architecture with a semantic latent space, and test it on DDM and StyleGAN2-ADA. Our results show that StyleGAN's W space is the most aligned with human notions of color in terms of vector similarities and generated colorways. Finally, we suggest that disentanglement can solicit a creative system for colorway creation, and evaluate it through expert questionnaires and within the lens of creativity theory.
\end{abstract}

\section{Introduction}

Color profoundly impacts the general sentiment expressed by a textile, influencing its perceptual structure and depth. Its importance is mirrored in the design brief preceding everyday clothing and furniture production. Alongside target specifications, technical requirements, and suggested moods, a crucial part of this brief indicates the desired \textit{colorways}~\cite{briggs2011textile, dickinson2011use}, alternate color variations of the same pattern. The market typically demands around 6-8 colorways to meet diverse societal needs~\cite{briggs2011textile}. 

\begin{figure*}[h]
  \centering
  \includegraphics[width=0.4\textwidth]
  {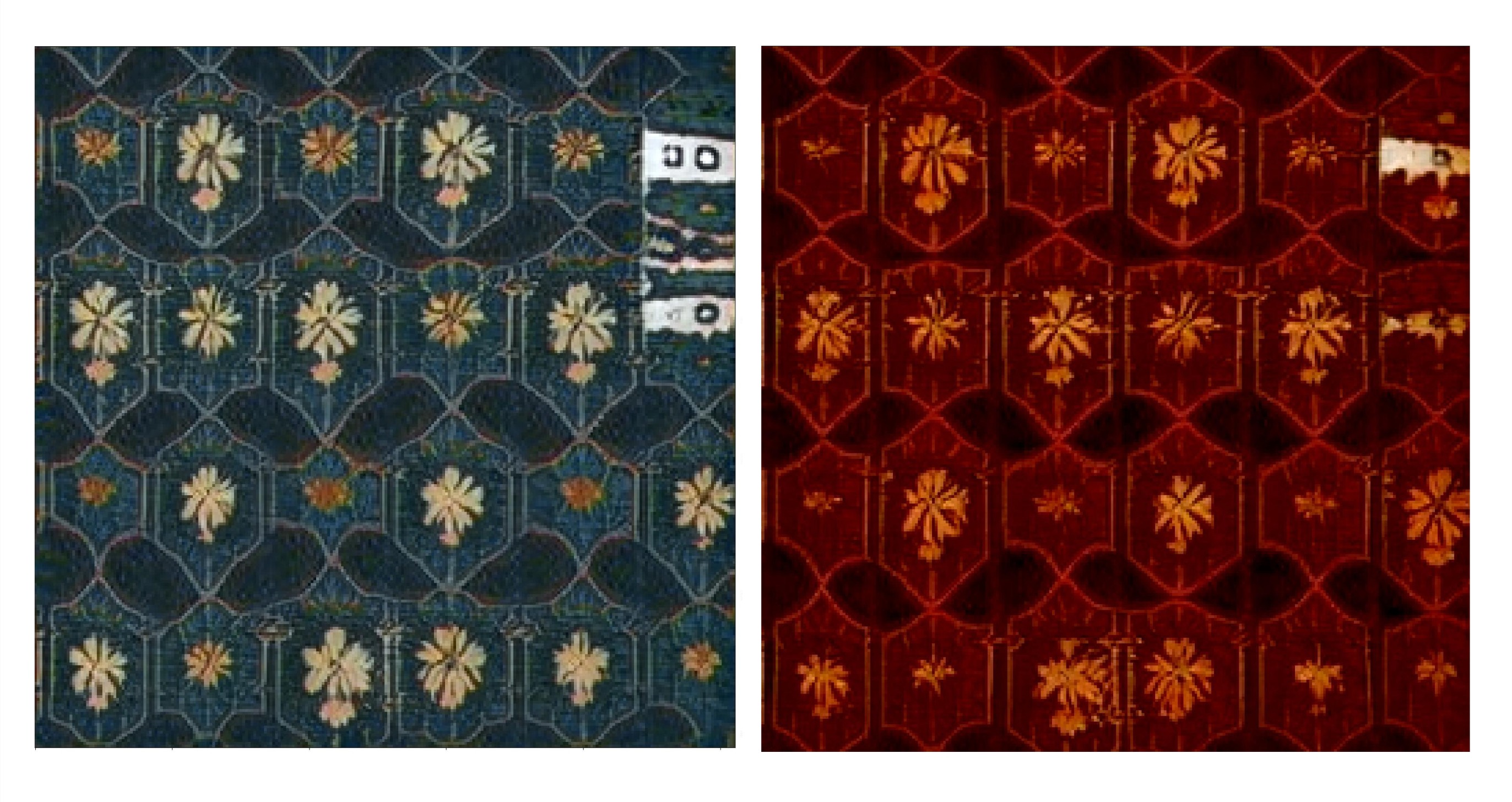}
  \caption{Example of a colorway in cold and warm tones. Pattern generated using StyleGAN2-ADA and colorway generated using InterfaceGAN.}
  \label{fig:colorway_example}
\end{figure*}

These colorways can range from slight base-color variations that maintain the original tone to completely different colorations that alter the feel, as in Figure~\ref{fig:colorway_example}. The creation of textile colorways is complex and subjective, influenced by historical and cultural contexts, technical discoveries, trends, seasonal requirements, and individual preferences~\cite{dickinson2011use}.

With traditional graphics programs and existing AI re-colorization methods, a color palette must be predetermined by the designer to change the color scheme of the graphic design. With ColorwAI, the framework proposed in the paper, designers and artists are assisted in the creative process, in the ideation of the colorway. The designer proposes a color and ColorwAI suggests the pattern in a novel color scheme and, so on, iteratively. In this way, the model prompts the designer to consider original possibilities, sometimes adjusting the design's shapes. The model, differently from re-colorization methods, adjusts the pattern slightly, to be more suitable to the new color scheme identified. It changes the shape of the pattern to the minimum, in a more harmonious way according to the knowledge gained from training on historical textiles. 

The purpose of the model is to suggest a "generative colorway", a colorway that generates both chromatic and slight pattern variation, to stimulate the designer's creativity. This is intended to accompany the creative process of the textile expert, who is traditionally given minimal specifications from the design brief and must present colorways that are stylistically and conceptually fitting the original design, but novel and original.  

ColorwAI tackles this creative process by relying on disentanglement. In deep neural networks, acquired representations are high-dimensional latent spaces encompassing a large variety of features. Disentanglement refers to the extraction of an interpretable feature such as color from the high dimensional latent space. Disentanglement learning is the task of learning a meaningful traversal direction in the latent space corresponding to one semantic variation of the output~\cite{choi2021not,bengio2013deep}. The latter is a well-established method for image editing and interpretation of the hidden representations of generative neural networks.

Because of its nature, weakly supervised disentanglement can determine the representation of color as a dimension within the latent space of generative neural networks~\cite{locatello2020weakly}. The color can be edited by perturbing along the latent direction, thus creating a "generative colorway" around that base color. Because generative neural networks such as VAEs, GANs, and Diffusion feature at least one coherently and semantically arranged latent space~\cite{karras2019style, kwon2022diffusion, higgins2016beta}, we assume that a disentangled color direction provides visually plausible color modifications, responding to the need for a suitable palette for the textile. Whenever a perturbation from a starting pattern along the latent direction of a color incurs into a lesser defined, warped, area of the latent space, the modifications are more substantial, incurring structural modifications. This may occur for a textile pattern when a certain chromaticity, e.g. blue, is lacking in the training data. In such cases, disentanglement creates slight pattern modifications, adjusting the original textile pattern into a similar one for which a blue example does exist in the training data. 

An additional advantage of disentanglement is that it allows understanding how colors are represented within these models in the context of textiles. Given the wealth of cultural, societal, and technical factors behind the use of color in textiles, we wish to complement the practical framework with insights into how our methods represent colors.

In this paper, we present ColorwAI, a framework for "generative colorway" creation that can be flexibly applied to any generative architecture with a semantic latent space. We identify four stages in the disentanglement pipeline: the coupling of latent codes with image annotations, the disentanglement of the latent space, the modifications of the latent code, and the inference with modified latent code. We show disentanglement results for StyleGAN and DDM~\cite{karras2019style, ho2020denoising}, maintaining distinct coupling and inference steps, but common disentanglement methods. This approach to Diffusion is a generalization of BoundaryDiffusion~\cite{zhu2024boundary}. We test InterfaceGAN and StyleSpace disentanglement methods and introduce a modification of InterfaceGAN, that we call \textit{ShapleyVec}. The latter uses a subset of latent space neurons for disentanglement~\cite{wu2021stylespace, shen2020interfacegan}. We demonstrate the superiority of the latter method in terms of quantitative performance and interpretability.

To summarize, the contribution of this work is as follows:
\begin{enumerate}
    \item We introduce a new task motivated by practical utility, the \textit{"generative colorway" creation}, that extends existing work on re-colorization with creative controls;
    \item  We present ColorwAI, a framework for color disentanglement that can be applied agnostically to all generative architectures presenting a semantic latent space, including GANs and Diffusion;
    \item  We enrich the methods that deal with disentanglement to favor better color disentanglement while maintaining interpretability and novelty through our modification of InterfaceGAN, \textit{ShapleyVec}. We interpret the color representations within these models.
    \item We justify the creativity of the framework in the context of Computational Creativity literature, conducting an expert survey.
\end{enumerate}

\section{Context}
\subsubsection{Textiles and Colors}

Colors and textures create complex interactions influencing human perception and artistic evolution. 
Textiles, originally referring to woven (from Latin \textit{textilis}) fabrics, have accompanied humans since the beginning of civilization, becoming a distinctive factor of humanity. In the West European-Mediterranean sphere, its history interlaced with that of color already 5000 years ago in Egypt~\cite{whiston1948design}. The colors came from the fibers and dyes. Their pigments were extracted from elements of nature, as in the ancient tradition of the madder root, creating shades of red and orange, and the expensive carmine-yielding cochineal insects. 

Pigment availability influenced cultural identities. Christian and Copts\footnote{The Copts are an indigenous Christian ethnoreligious community in Egypt.} colors are distinctively subdued in line with religious and social reasons, as opposed to the bright pigments of Assyrians and Babylonians, mirroring their active and strenuous civilizations but also the availability of the vibrant Tyrian purple~\cite{whiston1948design}. In more recent centuries, color developments instantiated trends that characterized their time, such as mauveine and analine purple in the 1850s and turquoise a century later~\cite{dickinson2011use}. 

Despite the current availability of plentiful synthetic dyes, colors continue to carry rich and layered meaning. 

The use of color in textiles is influenced by its perception. Studies pioneered by Chevreul's Laws of Contrast of Colour have shown that the co-location of colors into several schemes creates a harmonious product~\cite{burchett2002color}, becoming the de-facto standard in color design. Bauhaus artists such as Johannes Itten and Josef Albers, influenced by Goethe's Theory of Colors, explored color relationships affecting human perception. Josef Albers (1963) investigated color interaction. He demonstrated how perception of two identical colors on different backgrounds yields perceptually distinct colors~\cite{albers2013interaction}. Rules and costumes were introduced after the Great Exhibition in London of 1851, including the 'Grammar of Ornaments' by Owen Jones~\cite{jones2016grammar}. Jones mentions, in the 37 principles for textile design, the importance of using a pondered distribution of hues on the surface. 

Using digital synthetic textures, \cite{wang2020influence} demonstrates the influence of shape on the perception of the color composition in patterns, showing how more elongated shapes lead to less prominent chromatic impressions. The finding was already experimentally clear to Albers and Jones. Our perception of color similarity and arrangement is merely functional. Particularly, color patterns are very modestly understood by our vision compared to texture, contrast, and color~\cite{mojsilovic2000matching}.

\subsubsection{Texture Generation and Re-colorization}
Relevant research in Computer Vision has tackled the generation of textures~\cite{portilla2000parametric,karagoz2023textile, dai2014synthesizability, lin2023texture}. By texture synthesis, we mean the generation of images with structural components, spatially homogeneous and repeating, often fabrics or materials samples. Early works adopt parametric statistical models such as Portilla et al. \cite{portilla2000parametric}, using complex wavelet coefficients. More recently, papers have used deep learning-based generative models such as AE, GANs, and Diffusion. Bergmann et al. \cite{bergmann2018improving} introduce a structural similarity loss to detect textural defects using Autoencoders. Lin et al. \cite{lin2023texture} exploits StyleGAN to generate realistic visual textures, obtaining high-quality samples and inversion~\cite{karras2019style}. Lastly, Karagoz et al. \cite{karagoz2023textile} generate textile patterns using Stable Diffusion conditioned with a number of keywords from textile metadata~\cite{rombach2022high}.

Given the success of the last two models and the proven semantic arrangement of their latent space~\cite{karras2019style, kwon2022diffusion}, we generate textile samples and colorways using StyleGAN and Diffusion. 

The generation of colorways is closely linked with the task of image re-colorization. In this context, Xu et al. \cite{xu2019fabric} developed a re-colorization algorithm on fabrics based on parts segmentation and variation based on given color themes. Cho et al. \cite{cho2017palettenet} introduced the first deep neural network for content-aware image re-colorization. Yang et al. \cite{yang2021semantic}, among other factors, disentangle color mood within scene depictions. The color control is only partially addressed in \cite{yang2021semantic}. In the context of conditioning methods, Li et al. \cite{li2020manigan} use text conditioning to vary the color of semantic image segments using natural language. 

Re-colorization algorithms are time-consuming and constrained by requiring the selection of the predetermined palette, disentanglement methods are more flexible because they change the entire palette based on a coarse input, but have not comprehensively addressed the aspect of color. 

With ColorwAI, we wish to disentangle coarse color directions, that allow harmonious modifications based on a predominant guiding color. The choice of general controls is aimed at preserving the creative aspect of the framework. 

\subsubsection{Disentanglement}

Recent breakthroughs in deep neural networks for image generation have significantly impacted the artistic and design industries ~\cite{kingma2013auto,goodfellow2020generative,rombach2022high}, enhancing image editing and manipulation through disentanglement and text-conditioning techniques ~\cite{gatys2016image,karras2019style,patashnik2021styleclip,wang2023stylediffusion}. This study examines the creative and interpretative aspects of these advancements.

Methodologies for determining and interpreting the latent factors of variation have been developed starting from variations of VAE and GAN models in supervised and unsupervised settings~\cite{karras2019style,higgins2016beta}. Härkönen et al. \cite{harkonen2020ganspace}, for instance, discover interpretable image controls using Principal Component Analysis (PCA) on the latent space of the model. Using supervision, Shen et al. \cite{shen2020interfacegan} and Yang et al. \cite{yang2021semantic} train a machine learning regression to find the separation boundary between different semantics in the latent space. Methods such as Wu et al. \cite{wu2021stylespace} focus, instead, on the identification of specific dimensions and channels that disentangle the selected feature. 

Recently, attention has been dedicated to disentanglement in Diffusion models. This task is more complex than the disentanglement of earlier methods as Diffusion models do not present a unique deterministic latent space. Some works focus on the UNet bottleneck $h-space$ or latent representation $\epsilon-space$ using deterministic variations of the scheduler and inversion~\cite{ronneberger2015u,zhu2024boundary,song2020denoising}, while others disentangle the text conditioning~\cite{wu2023uncovering}.

\section{ColorwAI}

\begin{figure*}[h]
  \includegraphics[width=\textwidth]
  {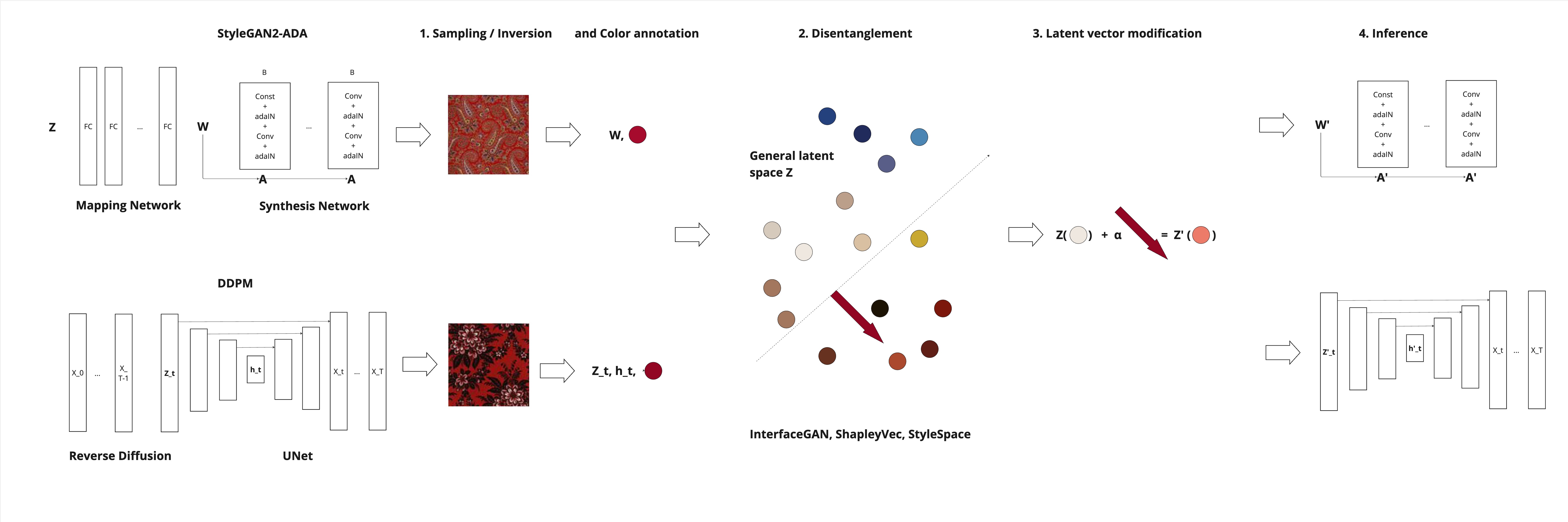}
  \caption{Schematization of the ColorwAI framework. We show the disentanglement steps for the two architectures, their latent spaces, and the inference procedure. Steps 1 and 4 are different for GANs and Diffusion, while 2 and 3 are common.}
  \label{fig:colorwAI}
\end{figure*}

This section introduces the "generative colorways" framework, detailing the \texttt{StyleGAN} and \texttt{DDM} models, their latent space properties, and disentanglement techniques. We describe methods for discovering latent traversal directions, including \textit{InterfaceGAN}, \textit{StyleSpace}, and our modified \textit{ShapleyVec} (Fig.~\ref{fig:colorwAI}).

\subsection{Latent Spaces}

\subsubsection{StyleGAN}
\texttt{StyleGAN2-ADA} is a style-based GAN generative neural network belonging to the family of StyleGANs~\cite{karras2019style}. The ADA version is more suitable for training with smaller data than the original and versions 2 and 3 of StyleGAN~\cite{karras2021alias}. 

The generative neural network comprises two parts: a mapping $F$ and a synthesis network $G$. The mapping $F: \mathcal{Z} \rightarrow \mathcal{W}$ receives as input a random code $\mathbf{z} \in \mathcal{Z}, \mathbf{z} \sim \mathcal{N}(0,1)$ and maps the vector through a series of fully connected layers to a latent vector $\mathbf{w} \in \mathcal{W}$. $\mathbf{w}$ is used by the synthesis network $G$ to generate the resulting image $\mathbf{x} = G(\mathbf{w}) \in  \mathcal{X}$. Both  $\mathcal{Z}$ and $\mathcal{W}$ are in $\mathcal{R}^{512}$. 

The mapping network generates a space, $\mathcal{W}$, that is more disentangled than the original latent space because it applies a non-linear transformation that helps separate and refine the individual factors of variation in the data.
This transformation allows $\mathcal{W}$ to achieve a more semantically meaningful representation where distinct features can be independently controlled and manipulated.

\subsubsection{Diffusion Denoising Models}
 In this section we outline the core principles of Diffusion models~\cite{ho2020denoising}, leaving in the appendix more details.

Diffusion models~\cite{ho2020denoising} generate data by learning how to reverse a forward diffusion process that gradually adds noise to the data. Starting from pure noise, the reverse process progressively refines the image to the desired outcome.

The forward process adds Gaussian noise to data points at each step:
\begin{equation}\label{ddpm}
q(\mathbf{x}t | \mathbf{x}{t-1}) = \mathcal{N}(\mathbf{x}t; \sqrt{1 - \beta_t} \mathbf{x}{t-1}, \beta_t \mathbf{I})
\end{equation}

The reverse process, parameterized by a neural network, aims to denoise these latent variables:
\begin{equation}
p_\theta(\mathbf{x}{t-1} | \mathbf{x}t) = \mathcal{N}(\mathbf{x}{t-1}; \mu\theta(\mathbf{x}t, t), \Sigma\theta(\mathbf{x}_t, t))
\end{equation}

DiffusionCLIP integrates diffusion models with CLIP to guide the generation process using text embeddings, by conditioning the reverse process~\cite{kim2022diffusionclip, radford2021learning}.

Asyrp propose to modify the generative reverse process to use the deepest feature maps~\cite{kwon2022diffusion}, the \texttt{h-space}, of a pre-trained diffusion model as semantic latent space for image manipulation, using a modified deterministic DDIM denoising schedule~\cite{song2020denoising}.

Boundary Diffusion segments the latent space into distinct regions for precise attribute control without additional parameters~\cite{zhu2023boundary}. This method modifies the reverse process to include boundaries during the mixing step, the denoising step where the final image starts to shape up. This method finds the disentangled direction of a feature by determining its semantic boundary using SVM. The latent direction is the orthonormal direction exiting this boundary, similar to the InterfaceGAN method used for GANs~\cite{shen2020interfacegan}. This indicates that other GAN disentanglement methods, such as StyleSpace and ShapleyVec, can also be applied within the Boundary Diffusion framework.

\subsection{Disentanglement Methods}

Given a well-trained model, disentanglement methods assume that the generated image $\mathbf{x}$ can be described by a set of semantics $\mathbf{s}$ and that there is a mapping from the $G: \mathcal{X} \rightarrow{} \mathcal{S}$. In particular, as we aim to generate colorways, we describe an image $\mathbf{x}$ by its predominant color, $\mathbf{s}$. Since there is a mapping between the latent space $\mathcal{Z}$ and the image space $\mathcal{X}$, there is also a mapping between the $\mathcal{Z}$ and the semantic space $\mathcal{S}$. 

In disentanglement, research has empirically shown that given two semantics, there is a hyperplane in the latent space that divides the two factors~\cite{shen2020interfacegan}. As such, there is a direction $\mathbf{n_s}$ in the latent space that manipulates a certain semantic. Given that direction, an image can be manipulated using $\mathbf{z_1} \rightarrow \mathbf{z} + \alpha \mathbf{n_s}$ to obtain a new image with that direction enhanced. In our case, if we generate an image $\mathbf{x}$ using $\mathbf{z}$ and we wish to make the image bluer, we need to find the direction that disentangles the factor blue $\mathbf{n_{blue}}$ and move positively towards that direction.

To find the traversal directions, the method of \texttt{InterfaceGAN} trains a linear SVM to find the hyperplane that separates each semantic and uses the orthonormal vector exiting the hyperplane as the disentangled direction $\mathbf{n}$~\cite{shen2020interfacegan}. Conversely, \texttt{StyleSpace} proposes a method to find the dimensions that deviate the most from the population mean, showing that the variation only present in those dimensions changes the desired semantic~\cite{wu2021stylespace}. 

\subsubsection{ShapleyVec}
A problem with \texttt{InterfaceGAN} is that the direction $\mathbf{n}$ found in latent space $\mathbf{z}$ utilizes all the dimensions of $\mathcal{Z}$ for the manipulation. For an ideal disentanglement, it should be possible both to manipulate the desired semantic without modifying any other aspect of the output (disentanglement) and to represent the semantic into only one dimension of the latent code (completeness). While the latter is often impossible, finding a direction that uses all dimensions will likely adopt nodes that do not encode that semantic, and therefore incur partial entanglements. Differently, dimensions that exhibit the largest variance, like in \texttt{StyleSpace}, may not take into account joint influences between dimensions and therefore do not provide a comprehensive direction.

In ShapleyVec, we build upon these two methods, maintaining their vantage points. First, we use a linear classifier like \texttt{InterfaceGAN} to find the optimal separation boundary between the two classes (images with a given primary color and images without that primary color). Second, we use Shapley values\footnote{The method was coined by Lloyd Shapley in 1953 as a method for collaborative game theory, that assigns payouts to players based on their contribution to the total payout. The method is widely used in machine learning interpretability, as it provides a quantitative and comparable value for feature importance.
} to analyze the contribution of each dimension to the final prediction~\cite{lundberg2017unified}. We filter out the dimensions that do not contain enough information about the desired feature. The filtering works as follows: we sort the dimensions by their Shapley values in descending order and maintain the first dimensions whose sum of the Shapley values reaches $E$ where $0 < E < 1$. Shapley values indicate the marginal contribution of each feature, with the total contribution summing to one. Using only the filtered dimensions, we train another linear classifier. We find the final direction $\mathbf{n}$ in a manifold space of $\mathcal{Z}$, where the dimensions that have been filtered out are masked using $0$. The orthonormal direction found using the second classifier is used. 

\section{Experimental Setup}
%\begin{figure*}[h]
 % \centering
  %\includegraphics[width=0.4\textwidth]{figures/Colorways - Example original image and generated images.jpg}
  %\caption{Piece of Russian Cotton. Late 19th century. Printed Textile. From Metropolitan Museum, Open Access.
%Left: Original image. Right: Image after background removal, inner mask square.}
 % \label{fig:sample_image}
%\end{figure*}
\subsubsection{Dataset and Processing}
We train our models on the textile collection of the IMET dataset from the Metropolitan Museum of Art, NY, including 20,000 textiles spanning from the sixth century BC to the 21st century across various civilizations. Additionally, we insert 10,000 textile samples from the Victoria and Albert Museum collections. After filtering out black-and-white images, we used 12,918 images for training, 4,324 for validation, and 4,318 for testing. To standardize the images, we remove backgrounds using the language-conditioned Segment Anything Mask~\cite{liu2023grounding, kirillov2023segany} with the text 'textile', extracting the largest spanning square within the mask. 

\subsubsection{Color Annotation}
To enable modifications based on a predominant guiding color, we annotate the main color of each image. Given a latent code $\mathbf{z}$ that generates an image $\mathbf{x}$, we annotate the color $\mathbf{s}$ of $\mathbf{x}$. We rely on CIELab and HSV color encodings, for their alignment with human vision and intuitive representation~\cite{mojsilovic2000matching}. We extract the color palette from each image, select a suitable color codebook, and quantize the main color in the palette. We empirically find that an adaptation of Bhatia's method~\cite{bhatia2004adaptive} using CIELab encoding provides the most pleasing and varied palette. We save the palettes and perform adaptive clustering to derive a set of $19$ predominant colors, forming the basis of the disentangled directions. Each image is assigned one of the codebook colors based on HSV similarity weighted $80\%$ on Hue, $10\%$ on Saturation, and $10\%$ on Value, to ensure that the annotations respect the original hue. For more information on the color annotation pipeline refer to appendices.

\subsubsection{Disentanglement Evaluation}
Inspired by the verification method developed by~\cite{ramesh2022hierarchical}, we evaluate the disentanglement of each of the color classes using a pseudo-accuracy score. We compare the performance of \texttt{ShapleyVec} to \texttt{InterfaceGAN} and \texttt{StyleSpace}, in \texttt{StyleGAN} and \texttt{DDM}.

The verification score is not suitable for multi-class disentanglement as there is no gradual increase to the desired semantic. Therefore, we first find the ideal intensity of modification $\alpha_{optimal}$ that allows changing the color of the image while maintaining most of its original structure. To do so, we test the disentanglement of each color at different $\alpha$ and register the \texttt{SSIM} (Structural Similarity Index Measure) to the original image. We find the optimal alpha for each $\mathbf{n}$ as:
\begin{equation}
    \alpha_{optimal} = \frac{1}{n} \sum_{i=1}^{N} \alpha*
\end{equation}

where $N$ is the number of samples and $\alpha*$ is:
\begin{equation}
\alpha* = max_{\alpha} [SSIM(\mathbf{z_0},\mathbf{z_{\alpha}}) \geq 0.75 \times SSIM(\mathbf{z_0},\mathbf{z_0})]
\end{equation}

$\mathbf{z_0}$ is the original image, and $\mathbf{z_{\alpha}}$ is the image modified in a direction by a factor $\alpha$. We choose to use the \texttt{SSIM} as it represents the perceived structural modification.

Once we find $\alpha_{optimal}$, we sample $m$ latent codes at random and annotate their main color $\mathbf{s}$. If an image is already in a given color, we remove it from the evaluation of that color. We then compute the pseudo accuracy ($p-acc$) on the remaining images. We calculate the percentage of images $\mathbf{x}$ generated by moving in the direction $\mathbf{n_c}$ of color $c$ by $\alpha_{optimal}$ whose main color is $c$. To take into account that some colors are more similar to others in the color wheel, we also propose a relaxed version of the accuracy ($relaxed-acc$), where the modification is $1$ if the final hue sits within the range of the desired color $c$, and $max(0, 1 - abs(d/100))$ where $d$ is the distance to the closest border.

\subsection{Experiments}
To find the disentangled direction $\mathbf{n_c}$ for each color $c$ in GANs, we annotate $1000$ latent codes $\mathbf{w} \in \mathcal{W}$ which we sample randomly from $\mathcal{N}(0,1)$. We focus on the latent space of \texttt{StyleGAN2-ADA} $\mathcal{W}$. For DDM, we invert $1000$ random images, which were previously annotated, using deterministic DDIM. We focus on the latent spaces $\mathcal{Z_t}$, the UNet input space, and $\mathcal{h_t}$, the UNet bottleneck layer, at mixing timestep $t=350$. We experiment with $100 < t < 500$ and empirically find $350$ to be the best tradeoff between adherence to the original pattern and color variation.

\subsubsection{Quantitative Results}
We test the performance of \texttt{InterfaceGAN} using several hyperparameters. We train both Logistic Regression (LR) and Support Vector Machines (SVM), the former with different regularization, $C \in [0.1, 0.01, 0.001]$. For \texttt{StyleSpace} we test extracting $10$, $20$, or $40$ dimensions. Lastly, for \texttt{ShapleyVec}, we assess if maintaining $0.25$ and $0.50$ of the explanation $E$ yields better results. We evaluate all methods using both accuracy $p-acc$ and relaxed accuracy $relaxed-acc$ sampling $100$ latent codes $\mathbf{w}$. We find that generally, LR yields slightly superior results, with $C=0.1$, $40$ dimensions for StyleSpace, and $E=0.5$ for ShapleyVec.

\begin{table}[h!]
    \centering
    \begin{tabular}{llll}
        \toprule
        Model&Method& p-acc. (std)&relaxed-acc. (std)\\
        \midrule
        StyleGAN&InterfaceGAN& 0.243 (0.14)&0.86 (0.08)\\
        StyleGAN&ShapleyVec& \textbf{0.287} (0.18)&0.89 (0.09)\\
        StyleGAN&StyleSpace& 0.20 (0.177)&0.874 (0.11)\\
 DDM& InterfaceGAN& 0.253 (0.41) & \textbf{0.901} (0.15) \\
 DDM& ShapleyVec& 0.194 (0.36) & 0.883 (0.17)\\
 DDM& StyleSpace& 0.247 (0.43) & 0.89 (0.16)\\
    \end{tabular}
    \caption{Results overview.}
    \label{tab:results}
\end{table}

The results of the evaluation are shown in Table~\ref{tab:results}. We observe that the standard deviations are generally high due to the great variability in performance among colors, especially for DDM. Moreover, we see a slight superiority of both \texttt{ShapleyVec} and \texttt{InterFaceGAN} to \texttt{StyleSpace} on StyleGAN. This is to be expected for GANs as \texttt{StyleSpace} is mostly intended to be used on the higher dimensional $\mathcal{W}+$ or $\mathcal{S}$ space of StyleGAN. Furthermore, \texttt{ShapleyVec} marginally surpasses \texttt{InterFaceGAN} in terms of accuracy and relaxed evaluation in StyleGAN, and vice-versa for DDM. Particularly, \texttt{ShapleyVec} has the best overall p-accuracy, obtaining more precise modifications, \texttt{InterfaceGAN} achieves the best relaxed accuracy, better approximating a general color mood.

\begin{figure*}[h]
  \includegraphics[width=\textwidth]{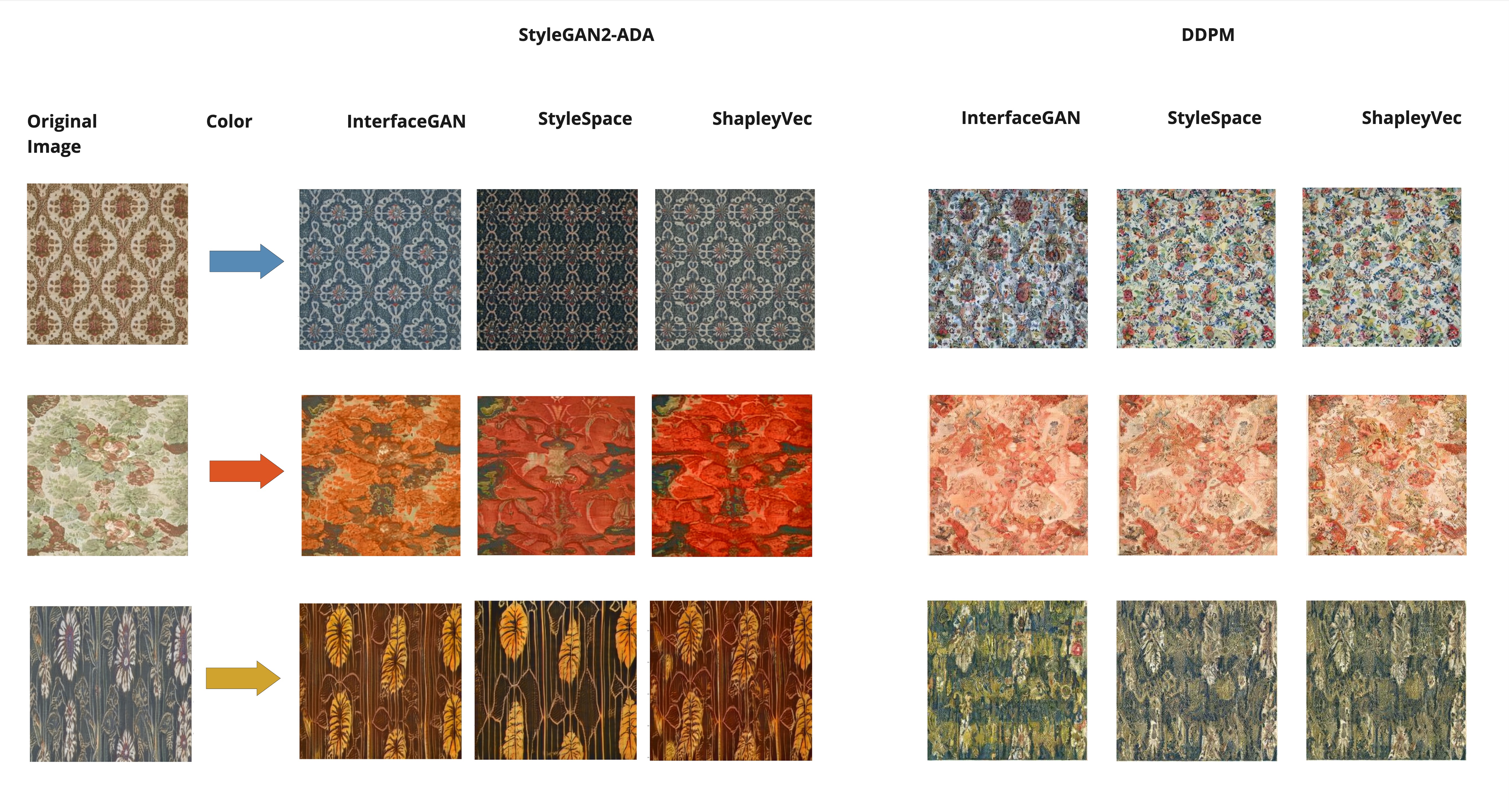}
  \caption{Colorways in three colors for three patterns using InterfaceGAN, ShapleyVec, and StyleSpace on StyleGAN and DDM. The examples are chosen because of their representativity of general observed trends.}
  \label{fig:colorways}
\end{figure*}

Some visual examples are shown in Figure~\ref{fig:colorways}. The first observation common to both architectures is that colors are not fully disentangled, and yield minimal shape modifications. While this is a downside in technical terms, the modifications enhance the creative potential of the system. Furthermore, we observe an astonishing difference between the type of color modification achieved by StyleGAN and by DDM. The first features more vivid shape and chromatic modifications. The shift to a neighboring pattern, visible in the first and second patterns from the top, appears seamless and visually pleasant, while the attempted transition in the last image of DDM towards yellow stripes manifests as drastic and incomplete. Furthermore, DDM maintains more precisely the shape of the original pattern, inserting chromatic modifications in the details. This difference may be due to the different setting of the disentanglement: while the modification in StyleGAN takes place prior to any decoding, the modified latent space is inserted in Diffusion only at the mixing step, where the skeleton of the image is already visible.

The examples show that InterfaceGAN is able to change the color with the smallest shape alterations. In the context of StyleGAN, however, the tonality of the final image is most faithfully achieved by ShapleyVec. For DDM, we observe a subtle small variation across methods. 

\subsubsection{Color Representation}
\begin{figure*}[h]
  \includegraphics[width=\textwidth]{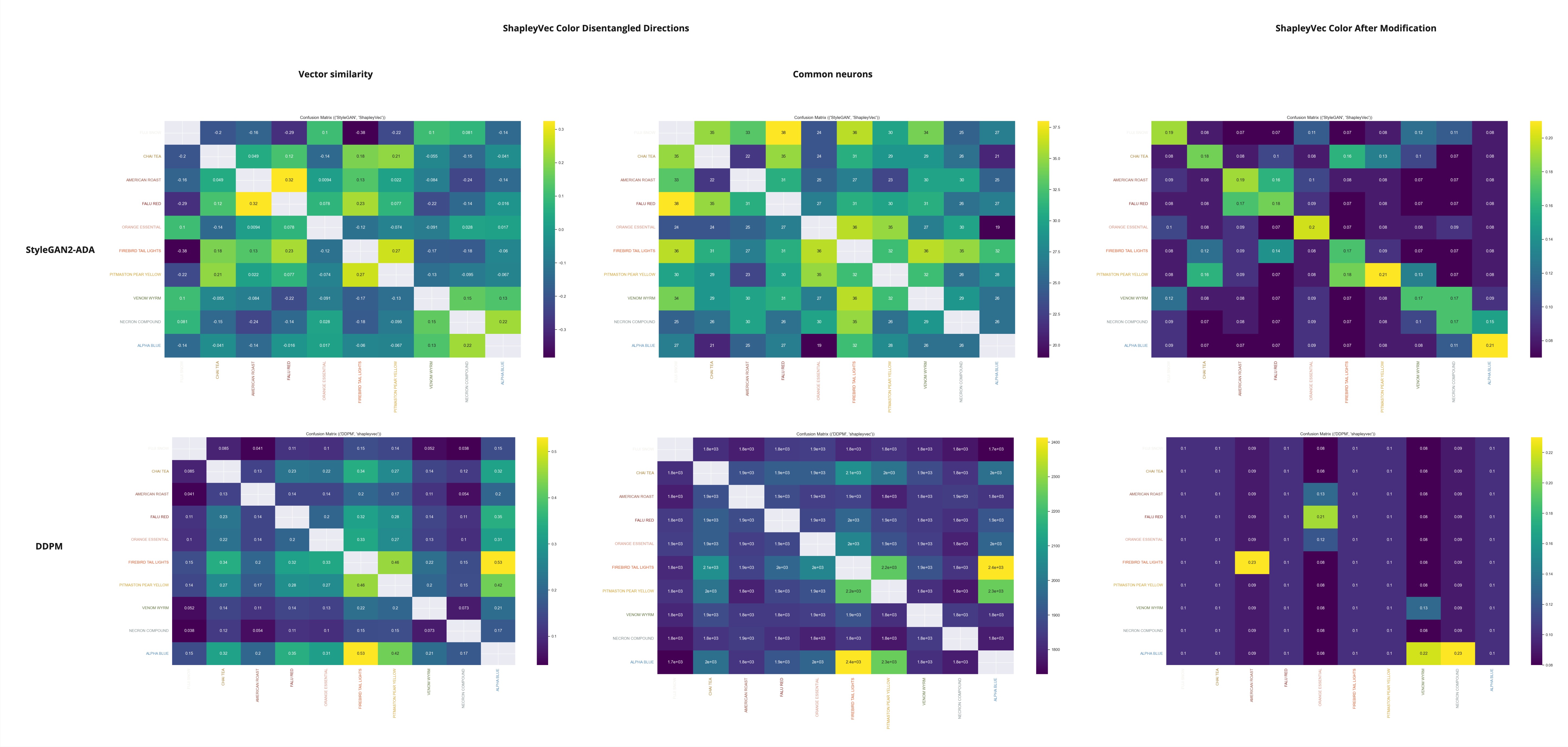}
  \caption{Relationship between color representations for StyleGAN and DDM using ShapleyVec. The relationships are shown in terms of similarity between the disentangled color vectors, in terms of which dimensions of the latent space encode which color(s), and of how often the modifications using such color directions result in other colors. For clarity, we only show results for a subset of color directions, which cover the spectrum of colors available, sacrificing some shades of brown and blue. }
  \label{fig:confusion_matrices}
\end{figure*}

An advantage of disentanglement is the creation of an explicit feature representation within the latent space of a model for each semantic. We leverage this property to explore the representation of each color and its features. We also show the confusion matrix of the disentangled directions following the quantitative evaluation. These experiments are displayed in Figure~\ref{fig:confusion_matrices}. This highlights a further advantage of ShapleyVec, for which we can inspect representations in terms of vector similarities and the subset of neurons of the latent space.

The leftmost column pictures the mutual cosine similarities between the feature representations. We observe that the representations within the $W$ space of StyleGAN broadly mirror our perceptual color similarity. In fact, beiges appear close to yellows, in turn, similar to oranges and themselves to browns. Similarly, blues and green are closely related. On the other side, the similarities within DDM's $h-space$ are not naturalistic, with the largest similarity appearing between Alpha Blue and orange-yellow tones. 

The central column figures which neurons in the latent spaces $W$ and $h-space$ are used in the color representations, and the overlap of the neurons across all pairs of colors. Like DDM's vectorial representation, the overlaps between colors are not naturalistic. Particularly, the number of overlapping neurons in DDM mirrors the vector similarities, due to the sparsity of the vectors. In fact, DDM's $h-space$ has 18'000 neurons, out of which ShapleyVec uses around 5000 to encode each color. In StyleGAN, we find two colors, Fuji Snow, a white, and Firebird Tail Lights, an orange, that have many common nodes to all other colors, as if they were more general 'color' representations. In this context, the white could simulate with a black-and-white direction and the orange with a saturation direction.

Lastly, the rightmost column shows the confusion matrices of the modifications using StyleGAN and DDM. While the confusion in StyleGAN is mostly with neighboring colors in digital spaces, DDM displays a clear preference towards mutations to certain colors.

Overall, it appears that the representations within StyleGAN are coherent with human representations, due both to the lower dimensionality of the space ($512$ vs $18'000$) and the uniqueness of the space as opposed to the denoising timesteps and skip connections present in diffusion.

\section{Discussion on Computational Creativity}\label{sec:discussion}

\subsubsection{Evaluation}
The premise of this framework is to create a stimulating interaction that boosts creativity. While the quantitative evaluation addresses the quality of the re-colorization, as discussed by Anna Jordanous, SOTA comparisons should accompany but not substitute qualitative evaluation~\cite{jordanous2022should}. The field of Computational Creativity equips us with tools to better frame this question~\cite{ritchie2019evaluation}. Using Ritchie's terminology, we can distinguish a creative system in a loose sense, carrying out activities considered by society as generally 'creative', and creativity in a strict sense, an attribute of a specific agent, action, or outcome based on achievement of a certain quality. The colorway creation, pertaining to the design sphere, can be coarsely considered as $creative_L$, in the loose sense. What we wish to assess is whether its outputs can be deemed $creative_S$, strict sense. To better understand the creativity of the framework, we conducted an expert qualitative evaluation of the framework and its results. This is referred to in the literature as an external evaluation with a symptomatic approach~\cite{ritchie2019evaluation}. We consider creativity in the eye of the beholder and inquire into which symptoms of creativity an expert recognizes or not in the system. 

The questionnaire is framed within Boden's creativity evaluation axes: novelty, surprise, acceptability, and quality~\cite{boden2004creative}. We assess novelty in P-creativity terms - novelty for the user in question; and surprise in terms of combinatorial, exploratory, and transformational creativity, further explained later~\cite{boden2004creative, ritchie2019evaluation}. To assess the quality we focus on the aesthetic value of the output and, lastly, we evaluate acceptability with respect to how much the system can be used in practice and, therefore, 'accepted' by the community. Details on the full questionnaire can be found in appendix.

The questionnaire was filled out by 4 design and/or textile experts and contains 5 sections with open-ended questions accompanied by an indicative 5-point Likert scale question. The results indicate that designers find the color variations mildly novel, with cases where the color is completely flipped being considered the most innovative. The results were mostly not indicated as surprising. The acceptability, on the other hand, was strongly favored, with the major drawback being that the material creation process is completely separated from the motif, unlike in real-world textile design. The quality was also very positively evaluated, and the aesthetics of the color modifications were mostly appreciated for the sharpness and contrast of the result. The major drawback is the seldom loss of contrast between colors, which makes the shapes less legible.

Interestingly, the evaluators all displayed a preference for contemporary designs, suggesting a preference for abstract shapes, and saturated and contrasting colors. This highlights a major limitation of the current approach, which is trained on historical textiles and uses nude tones as disentangled directions. The perception of limited novelty and surprise is also influenced by the historical style of the results.

\subsubsection{Type of Creativity}

Margaret Boden~\cite{boden2010creativity} identifies three forms of creativity: combinatorial, exploratory, and transformational creativity. The first is the ability to generate new elements through the combination of existing ones; the second is the generation of new forms through the exploration of conceptual spaces, while the last refers to an abstract transformation.

Disentangled semantic directions allow exploring the potential states of the model, the colors that are 'plausible' for certain textiles. We expect the model to change a design to a color only to the extent that it makes sense within its 'conceptual' representation of the data. Following this reasoning, the disentangled color directions, which we originally claimed to stimulate creativity, would allow \textit{exploratory creativity}, as they would enable using a basic concept to explore modifications that go beyond that concept itself and into the cultural framework of the model. 

Within this type of creativity, historicity becomes a value, a framework aimed at being surpassed. The framework allows generating characteristic examples and mutating the colors according to what has been historically seen, whilst allowing for variations. The addition of novel colors, shapes, and other forms of art can further strengthen the ability to learn from the rule and go beyond it. In this sense, following a direction of the latent space posits this type of generation in an advanced stage of Dan Ventura's Mere Generation steps~\cite{ventura2016mere}. In fact, such a process can be considered a filtering step using visual perception, leaving only the inception of a knowledge base as the last missing step towards creativity and away from the critiqued 'mere generation'.

\section{Conclusion}
In conclusion, we have explored the task of "generative colorway" creation using disentanglement learning. We proposed ShapleyVec, which marginally outperforms other disentanglement methods on StyleGAN and allows a more complete interpretation of the color representation in three ways. We find StyleGAN's $W$ space to be most aligned with human color perception and to yield the most seamless modifications. We compare the results with disentanglement on DDM, establishing a shared framework, and identifying a key visual difference in the edits. While StyleGAN changes 'semantically' the output, DDM disentanglement based on the mixing step only allows to 'redirect' the output toward the new color. Finally, we suggest that disentanglement can serve as a method to leverage the exploratory creativity of latent spaces using Margaret Boden's terminology. 

We acknowledge the limitations of this paper which open avenues for future research. Further tests and evaluations should be carried out to be able to assert the superiority ShapleyVec over InterfaceGAN, including a quantitative evaluation for creativity and testing on more samples. To enrich the comparison between GAN and Diffusion, more disentanglement methods on DDM not using the mixing step, such as Asyrp, should be tested. Lastly, to enrich the novelty and surprise of the output, evolutionary art methods, more disentangled factors and data from a larger temporal spectrum can be included in future work.

\section*{Acknowledgments}
Digital Visual Studies is a project funded by the Max Planck Society. We would like to thank Pepe Ballesteros Zapata and Dario Negueruela del Castillo for the continuous exchange of ideas, Fabrizio Silvestri for the support and for creating contact between the authors; and Valentine Bernasconi and Davide Torre for the invaluable feedback on the manuscript.

\bibliographystyle{splncs04}
\bibliography{main}

\appendix

\clearpage
\setcounter{page}{1}
\setcounter{section}{0}

\section{Diffusion Denoising Models}
Diffusion models \cite{ho2020denoising} generate data by simulating a reverse diffusion process. The core idea is to learn how to reverse a forward diffusion process that gradually adds noise to the data. Starting from pure noise, the reverse process progressively refines the image to the desired outcome.

Let $\mathbf{x}_0$ be a data point from the true data distribution. The forward process defines a sequence of latent variables $\{\mathbf{x}_t\}_{t=0}^T$ by adding Gaussian noise at each step:

\begin{equation}\label{ddpm}
q(\mathbf{x}_t | \mathbf{x}_{t-1}) = \mathcal{N}(\mathbf{x}_t; \sqrt{1 - \beta_t} \mathbf{x}_{t-1}, \beta_t \mathbf{I})
\end{equation}

where $\beta_t$ are the variance schedule parameters controlling the amount of noise added at each step. The reverse process, parameterized by a neural network, aims to denoise the latent variables step by step:
\begin{equation}
p_\theta(\mathbf{x}_{t-1} | \mathbf{x}_t) = \mathcal{N}(\mathbf{x}_{t-1}; \mu_\theta(\mathbf{x}_t, t), \Sigma_\theta(\mathbf{x}_t, t))
\end{equation}

The neural network $\mu_\theta$ is trained to approximate the true reverse conditional distribution, allowing the model to progressively denoise from $\mathbf{x}_T$ (pure noise) back to $\mathbf{x}_0$ (data sample).

%\subsubsection{Semantic Latent Spaces}
Diffusion models are currently state-of-the-art in image generation but the presence of a suitable semantic latent space to control the generation process is still under discussion. They iteratively transform a random noise vector to produce a high-resolution image, but there is no single semantic mapping between the vector and the image’s features due to a lack of structured hierarchy in the models, making it difficult to edit specific attributes of the generated image.

\texttt{DiffusionCLIP} \cite{kim2022diffusionclip} integrates diffusion models with Contrastive Language-Image Pre-training (CLIP) \cite{radford2021learning} to guide the diffusion process, enabling the generation of images that are semantically aligned with provided textual descriptions. Let $\mathbf{y}$ be the textual description and $\mathbf{e}_y$ be the corresponding text embedding obtained from CLIP. The diffusion model's reverse process is then conditioned on this text embedding $\mathbf{e}_y$:
\begin{equation}
p_\theta(\mathbf{x}_{t-1} | \mathbf{x}_t, \mathbf{e}_y) = \mathcal{N}(\mathbf{x}_{t-1}; \mu_\theta(\mathbf{x}_t, t, \mathbf{e}_y), \Sigma_\theta(\mathbf{x}_t, t, \mathbf{e}_y))
\end{equation}

\texttt{Asyrp} \cite{kwon2022diffusion} propose to modify the generative reverse process to use the deepest feature maps, the \texttt{h-space}, of a pre-trained diffusion model as semantic latent space for image manipulation. DDIM \cite{song2020denoising} redefines Eq. \ref{ddpm} as 
\begin{equation}
 q_{\sigma}(x_{t-1}|x_t, x_0) = \mathcal{N} \left( \sqrt{\alpha_{t-1}} x_0 + \sqrt{1 - \alpha_{t-1} - \sigma_t^2} \frac{x_t - \sqrt{\alpha_t} x_0}{\sqrt{1 - \alpha_t}}, \sigma_t^2 \mathbf{I} \right)
\end{equation}

which can be rewritten as:
\begin{equation}\label{ddim}
x_{t-1} = \sqrt{\alpha_{t-1}} P_t(\epsilon_{\theta}(x_t | \Delta h_t)) + D_t + \sigma_t z_t,
\end{equation}

where $P_t$ is the predicted image, $D_t$ is the direction pointing to $x_t$, $\sigma_t$ is the variance of the reverse process, and $z_t \sim \mathcal{N}(0, I)$ is a noise term. In \texttt{Asyrp}, only $P_t$ is modified while $D_t$ remains unchanged. The shift $\Delta h_t$ is optimized to achieve the desired $P_t$  using a directional CLIP loss:

\begin{equation}
L_{\text{direction}}(x_{\text{edit}}, y_{\text{target}}; x_{\text{source}}, y_{\text{source}}) = 1 - \frac{\Delta I \cdot \Delta T}{\|\Delta I\|\|\Delta T\|},
\end{equation}

where $\Delta T = E_T(y_{\text{target}}) - E_T(y_{\text{source}})$ and $\Delta I = E_I(x_{\text{edit}}) - E_I(x_{\text{source}})$, with $E_T$ and $E_I$ being the CLIP text and image encoders, respectively.

\texttt{Boundary Diffusion} \cite{zhu2023boundary} enables precise attribute control by directly segmenting the latent space into distinct regions without adding any extra parameters to the base diffusion model. This is in contrast to Asyrp, which requires training auxiliary editing neural networks, and DiffusionCLIP, which fine-tunes the pre-trained diffusion model. Boundary Diffusion introduces a modified reverse process of eq. \ref{ddim}  to include boundaries in the latent space:
\begin{equation}
x_{t-1} = \sqrt{\alpha_{t-1}} P_t(\epsilon_{\theta}(x_t | B_t(\Delta h_t))) + D_t + \sigma_t z_t
\end{equation}

where $\epsilon_{\theta}$ is the noise predictor conditioned on the modified feature maps $h_t$, and $B_t(\Delta h_t)$ represents the boundary function applied to the shift in the latent variables. Furthermore, the boundary is introduced in the latent space during the mixing step, the denoising step at which the final image becomes clear.

BoundaryDiffusion finds the disentangled direction of a feature by obtaining the semantic boundary of that feature using SVM. The latent direction is the orthonormal direction exiting the semantic boundary. The method to find the disentangled direction is equivalent to the InterfaceGAN proposed for GANs~\cite{shen2020interfacegan}. The applicability of this method on the latent space of Diffusion indicates that, within the BoundaryDiffusion framework, also the other GAN disentanglement methods, StyleSpace and ShapleyVec, can be applied.

\section{Color annotation}
\subsubsection{Color Annotation}\label{sec:color}
As we wish to disentangle coarse color directions, that would allow harmonious modifications based on a predominant guiding color, we annotate the predominant color of each image. Given a latent code $\mathbf{z}$ that generates an image $\mathbf{x}$, we annotate the main color $\mathbf{s}$ of $\mathbf{x}$. Digitally, color can be represented in different encodings, notably, RGB, HSV, CIELab, XYZ, each enforcing an inherent concept of similarity and gamuts. Particularly, the encoding of CIELab is most aligned to human vision, while HSV (hue, saturation, value), is the most understandable and intuitive color representation~\cite{mojsilovic2000matching}.

\begin{figure*}[h]
  \centering
  \includegraphics[width=\textwidth]{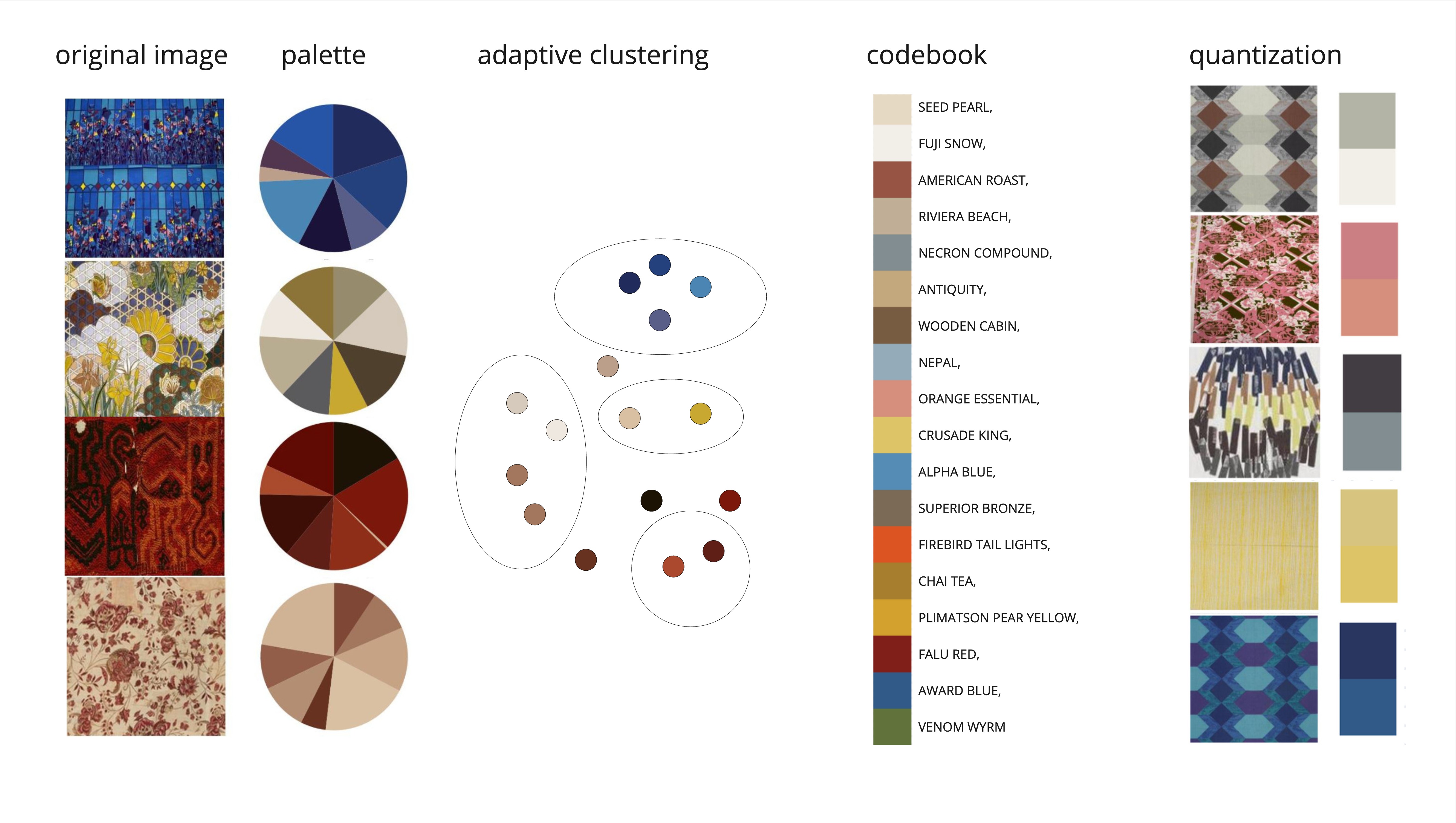}
  \caption{Predominant color annotation. From left to right: palette extraction, codebook identification, color quantization. The final annotated color is the lower box of each rightmost image.}
  \label{fig:color_annotation}
\end{figure*}

We wish to annotate an image with the color that most completely represents the perception it stimulates, while concurrently reducing the number of possible color directions to a sufficiently varied but manageable set. We tackle this in three steps: the color palette extraction of each image, the selection of a suitable color codebook for quantization, and the final quantization of the main color in the palette. The predominant color annotation process can be found in Figure~\ref{fig:color_annotation}.

Several methods have been developed to extract palettes from an image, often in the domains of image compression or retrieval. The most simple method relies on k-means to identify a set of $K$ clusters from the $RGB$ or $Lab$ colors of an image and take the cluster centroids to be the palette. While this method works well, it fails to extract contrasting colors, which stand out the most to the human eye. Modifications have been proposed, including adaptive clustering methods~\cite{pappasadaptive, bhatia2004adaptive}, and codebook-based quantization methods~\cite{mojsilovic2000matching}. We test the different methods on our data and find empirically that an adaptation of \cite{bhatia2004adaptive}'s method using $CIELab$ encoding results in the most pleasing and varied color palette.

Successively, we save the extracted color palettes for all images in the training set and reiterate the adaptive clustering to obtain a data-driven set of predominant colors present in textiles. The resulting set of 19 colors becomes the basis of the disentangled directions. 

Lastly, to assign each image to one of the codebook colors, we retain the first color of the palette and quantize it based predominantly on the Hue similarity. In fact, as we wish to ensure that the annotations respect the hue of the original image, and are only partly concerned with saturation and brightness, we base encode the colors in $HSV$ and weigh each channel based on empirical performance. We create a similarity metric using $80\%$ Hue, $10\%$ Saturation, and $10\%$ Value. 

\section{Questionnaire}
This appendix documents the questions and visuals of the expert questionnaire.

\begin{enumerate}
\item \textbf{Introduction}
  
    \begin{itemize}
      \item Name
      \item Email
      \item Job position and company/institution
      \item On a scale from 1 to 5, how do you evaluate your level of experience with design and textiles? (1 not expert, 5 expert)
    \end{itemize}

\item \textbf{Creativity}

    \begin{itemize}
      \item Novelty. On a scale of 1 to 5, how innovative/original do you find the color combinations generated compared to existing designs? (1 not new at all, 5 extremely new) 
      \item If you think one or more combinations are innovative, indicate which ones and explain why
      \item Surprise. On a scale of 1 to 5, how surprising do you rate the color and pattern combinations generated? (1 not at all surprising, 5 extremely surprising)
      \item Based on the examples presented, how would you define the type of surprise they elicit?
      \item Usefulness. On a scale of 1 to 5, how applicable are the examples of "generative colorway" to possible commercial textile collections? (1 not useful at all, 5 extremely useful)
      \item Why?
    \end{itemize}

\item \textbf{Usability}

    \begin{itemize}
      \item Ease of interaction. On a scale of 1 to 5, how do you find the choice of color modification based only on a predominant color and not on the entire palette? (1 very difficult, 5 very easy)
      \item What advantages and disadvantages do you identify? 
      \item Control and flexibility. On a scale of 1 to 5, do you feel that you can have sufficient control over the colors using this simplified model? (1 no control, 5 complete control)
      \item What improvements could be introduced to increase control and/or flexibility?
      \item Color choice. On a scale of 1 to 5, how comprehensive do you find the spectrum of colors available to you? (1 not exhaustive at all, 5 comprehensive)
      \item Assuming, for example, that you need to rework a historical textile, what color is missing? How many color choices would you like to have (there are currently 19)?
    \end{itemize}

\item \textbf{Aesthetics}

    \begin{itemize}
      \item Visual attractiveness. On a scale of 1 to 5, how aesthetically pleasing do you find the generated color combinations? (1 not pleasing at all, 5 extremely pleasing)
      \item What specific elements of color combinations contribute to greater appeal? 
      \item Shape modifications. On a scale of 1 to 5, how do slight pattern adjustments contribute to the overall aesthetic quality of the fabric? (1 detrimental, 5 significantly improving)
      \item Are there specific adjustments that you have found particularly effective or not?
      \item Color combination. On a scale of 1 to 5, how harmonious are the colors used in the textile? (1 not harmonious at all, 5 extremely harmonious)
      \item What color combinations do you like best and why?
    \end{itemize}

\item \textbf{Consistency with textile design practices and traditions}

    \begin{itemize}
      \item Consistency. On a scale of 1 to 5, do color combinations and patterns maintain consistency with established design principles? (1 not consistent at all, 5 highly consistent)
      \item In what examples, do you find that more or less of these principles have been adhered to? Which ones?
      \item Historical Pertinence. On a scale of 1 to 5, how consistent are the color palette combination and pattern with the historical period from which the pattern came? (1 not accurate at all, 5 highly accurate)
      \item Among the examples, are there any styles or historical periods with which the AI color combinations would be assonant or not?
    \end{itemize}
    
\item \textbf{Coherence between pattern and color}

    \begin{itemize}
      \item Necessity of the pattern change. On a scale of 1 to 5, how necessary do you think the pattern adjustments made by the AI to accommodate the new color scheme are? (1 not necessary at all, 5 extremely necessary)
      \item Explain why you think these adjustments are necessary or unnecessary.
      \item Improvement in consistency. On a scale of 1 to 5, how do these minimal changes in shape impact pattern-color consistency? (1 significant worsening, 5 significant improvement)
      \item Among the examples, are there instances where consistency was particularly strengthened or not?
    \end{itemize}

\item \textbf{General feedback and suggestions}
    
    \begin{itemize}
      \item General impression. What is your overall impression of ""generative colorway""? Are there any aspects that you particularly liked or disliked?
      \item Suggestions for improvement. Do you have suggestions for improving AI performance in ""generative colorway""? Are there additional features or controls that you would like to see integrated into the system?
    \end{itemize}

\end{enumerate}

\begin{figure*}[h]
  \centering
  \includegraphics[width=0.8\textwidth]{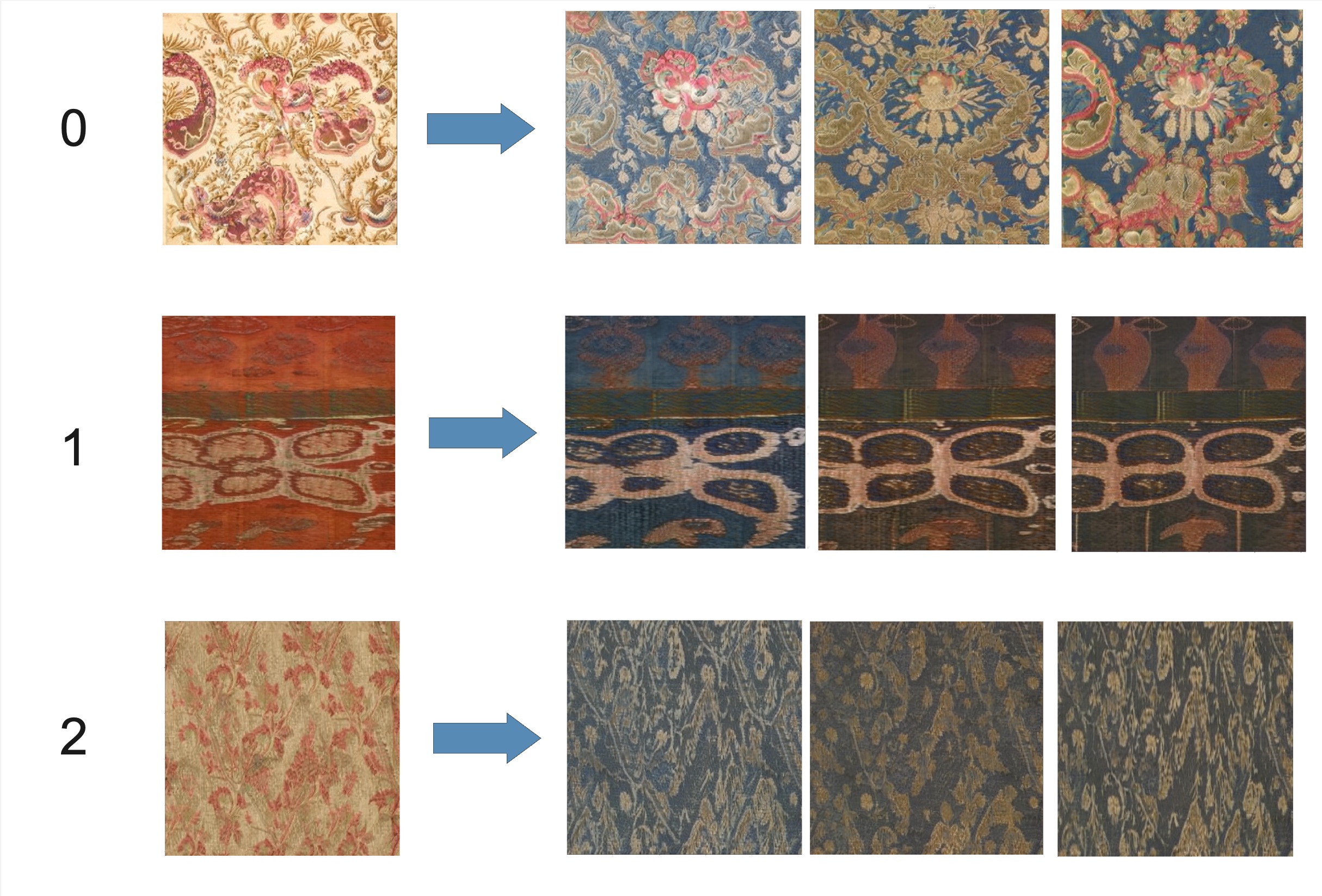}
  \includegraphics[width=0.8\textwidth]{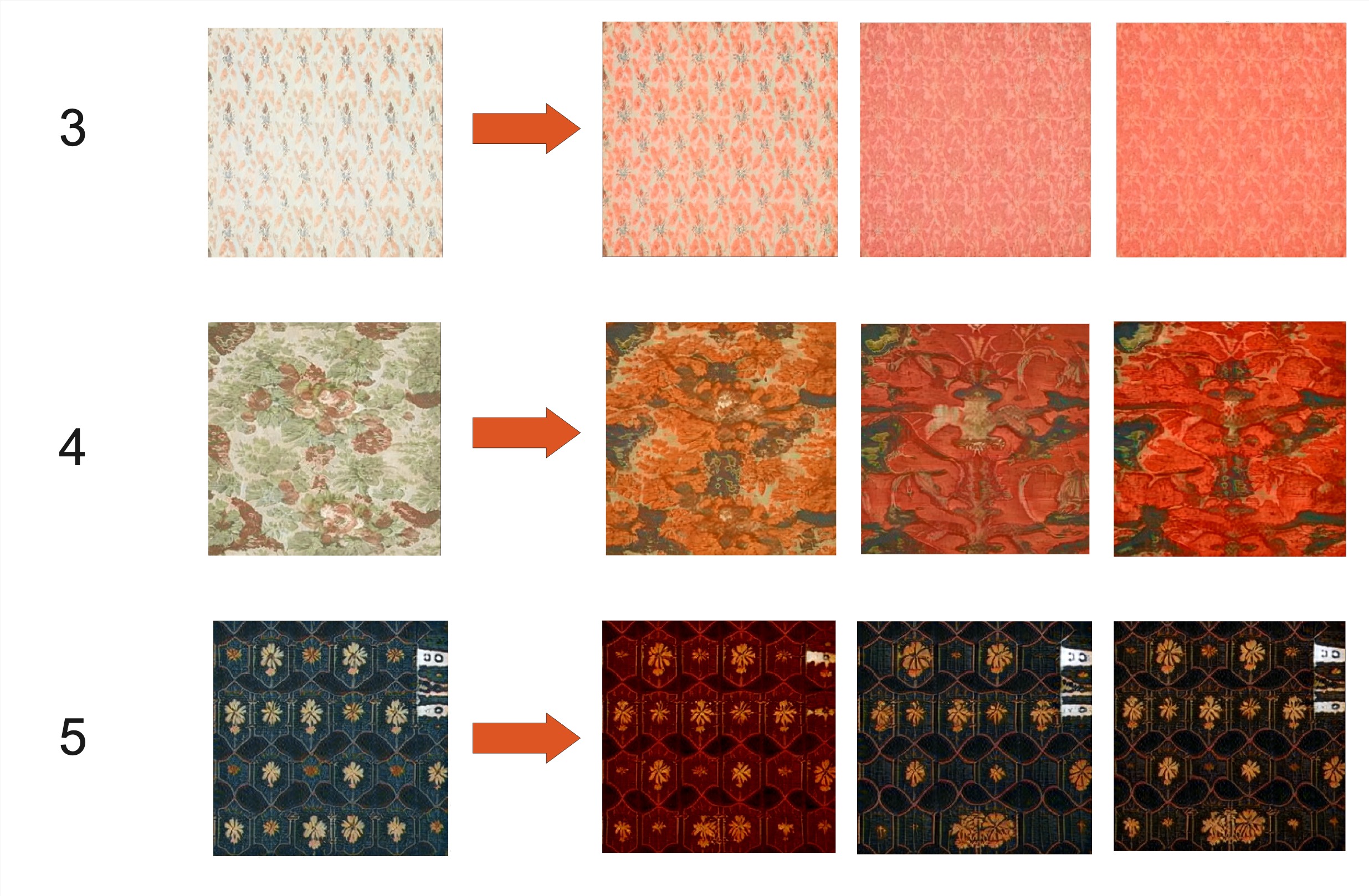}
  \includegraphics[width=0.84\textwidth]{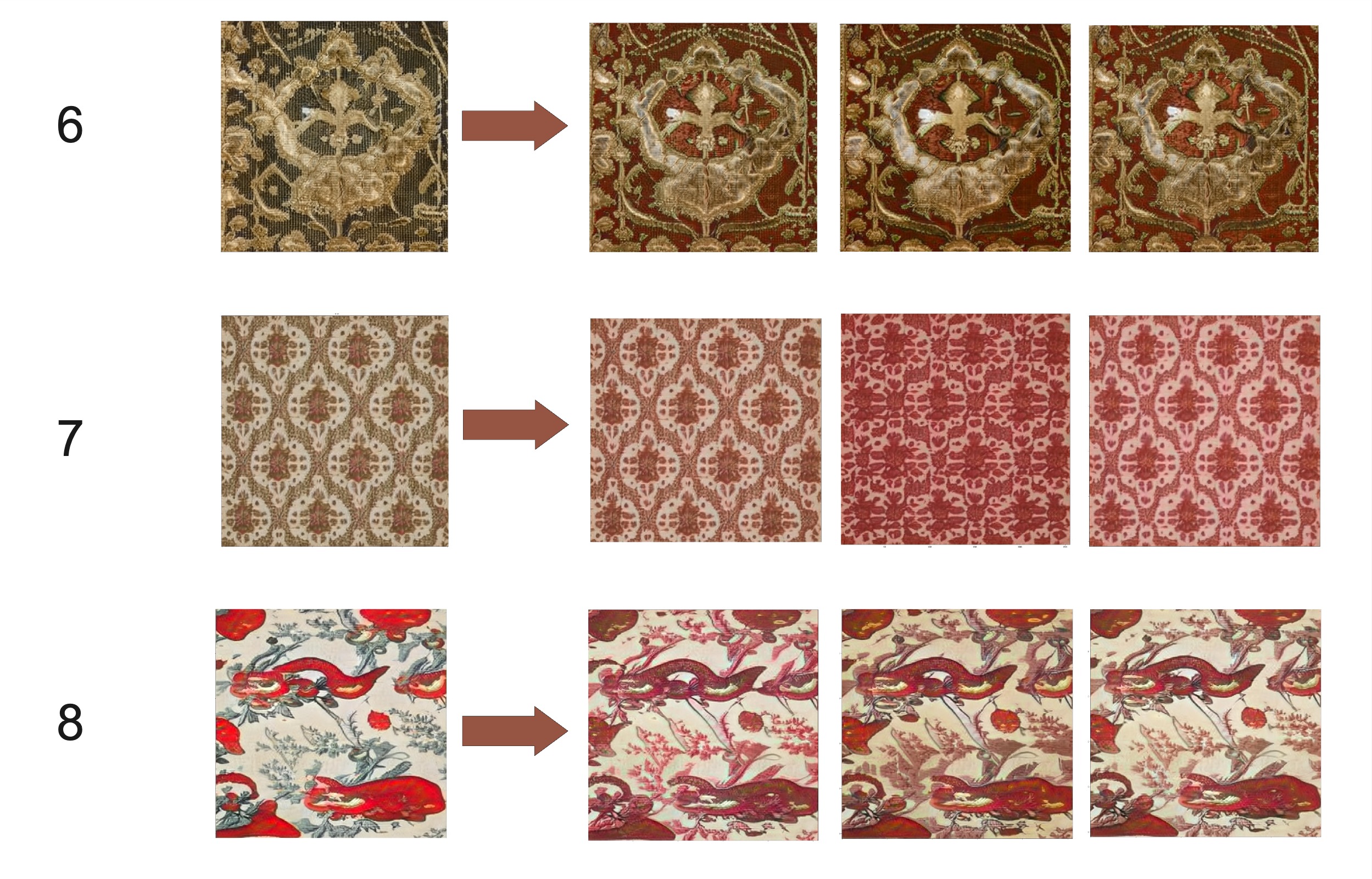}
 
  \label{fig:questionnaire}
\end{figure*}

\begin{figure*}[h]
  \centering
  \includegraphics[width=0.8\textwidth]{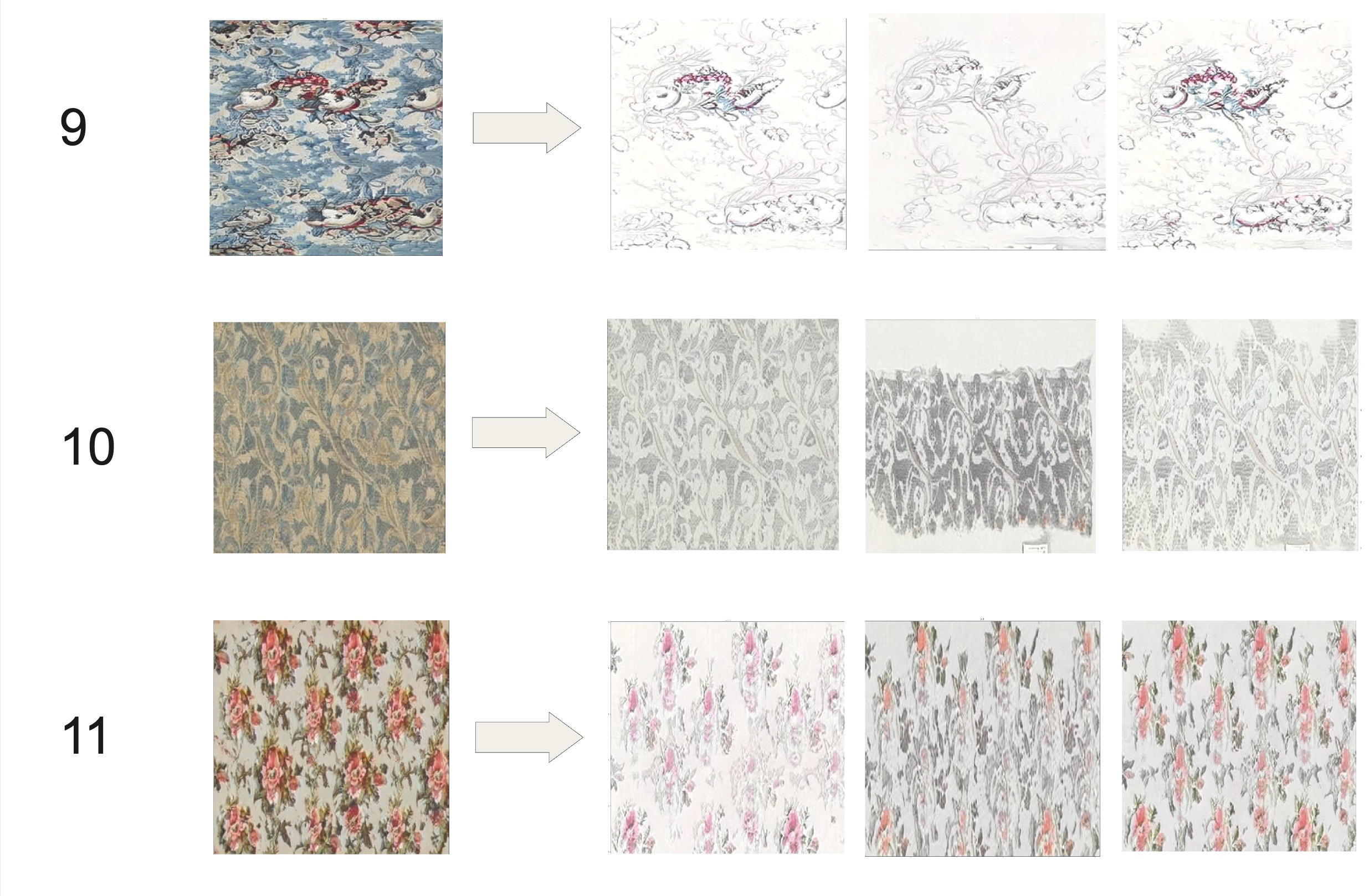}
  \includegraphics[width=0.8\textwidth]{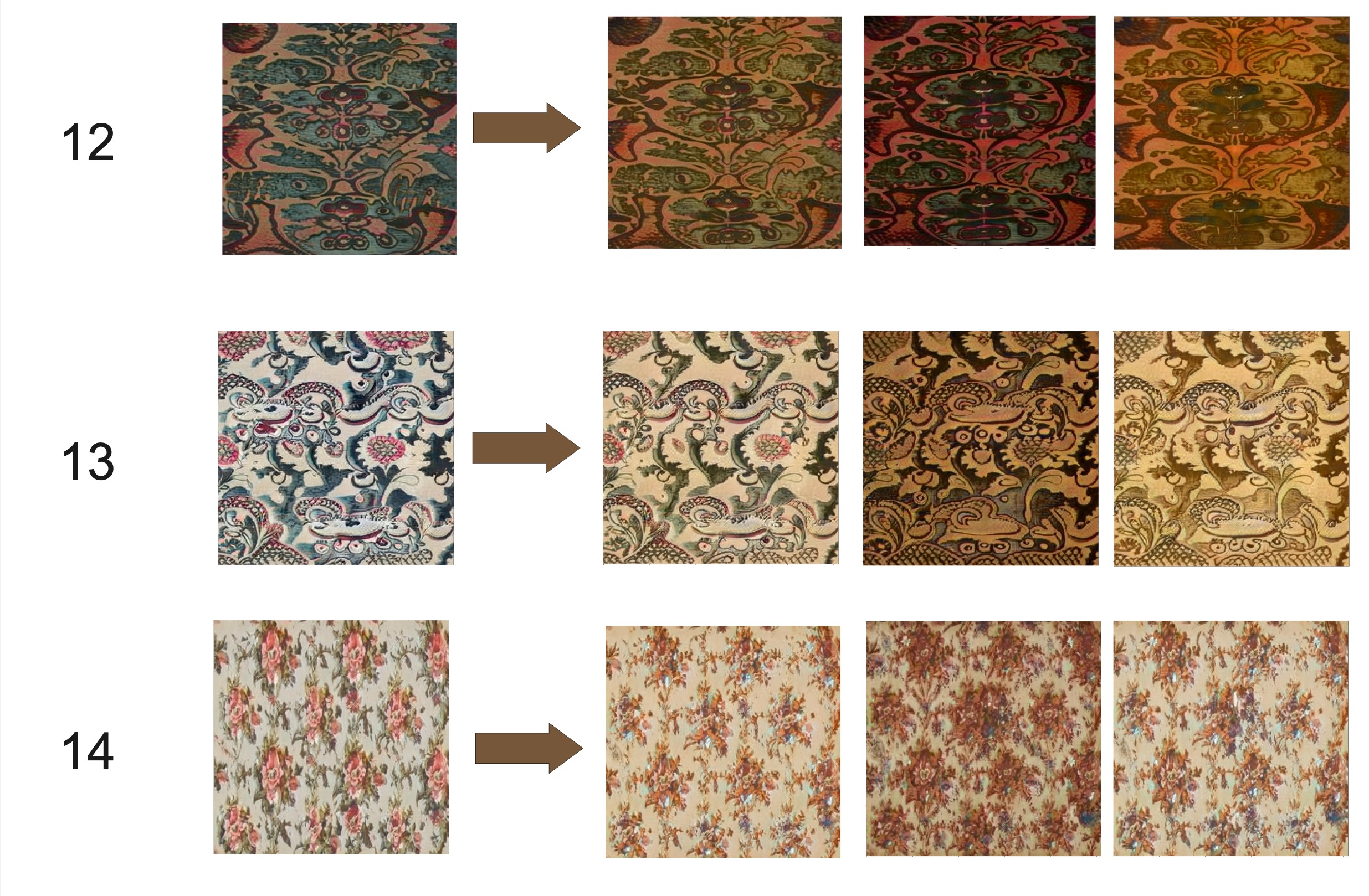}
  \includegraphics[width=0.8\textwidth]{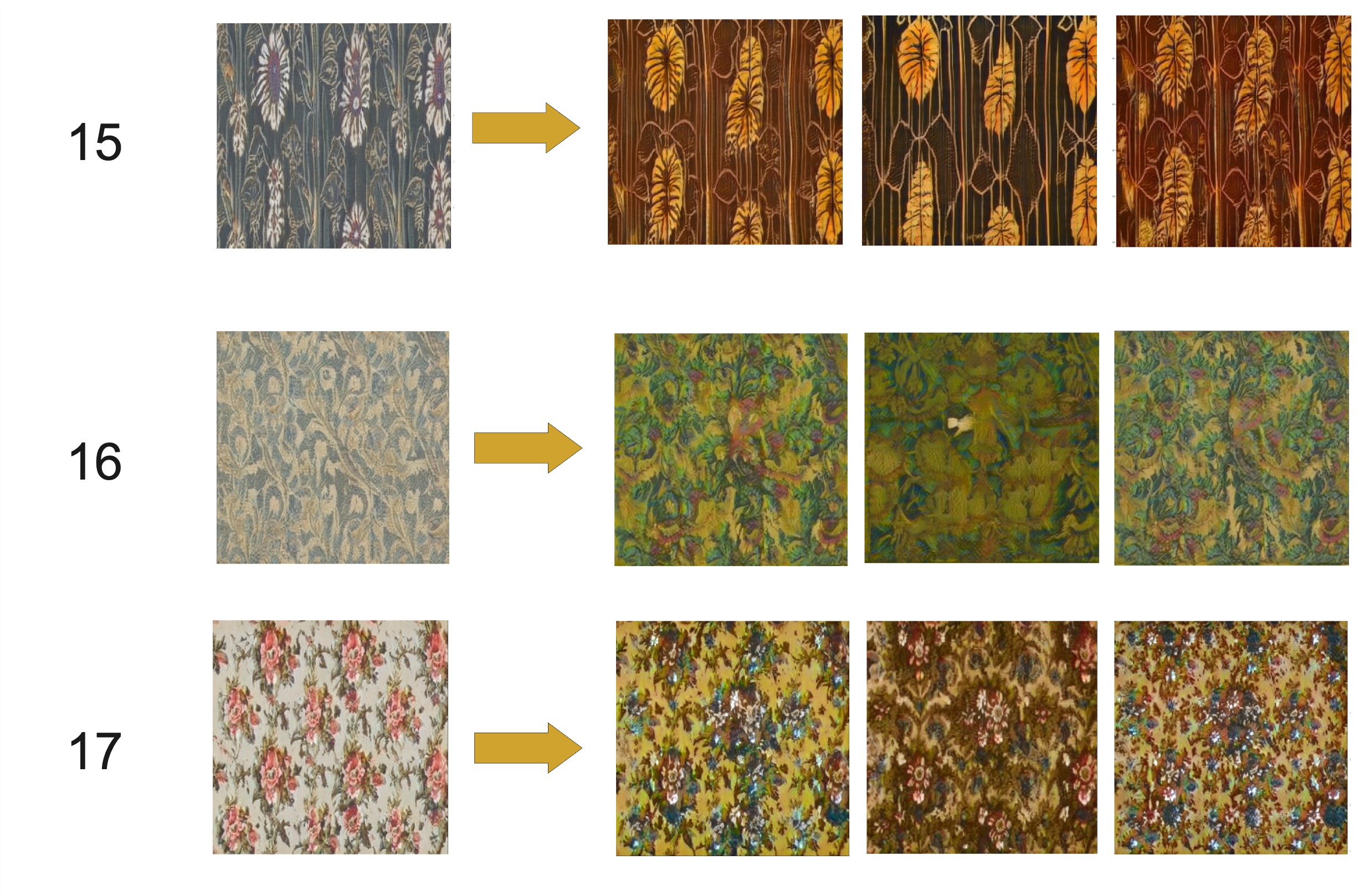}
  
  \caption{Examples of "generative colorways" to be considered in each section. On the left is the image of the original pattern; on the right are three versions of "generative colorway" in the guide color indicated by the arrow.}
  \label{fig:questionnaire2}
\end{figure*}
    
\end{document}